\documentclass{article}

\usepackage{arxiv}

\usepackage[utf8]{inputenc} 
\usepackage[T1]{fontenc}    
\usepackage{hyperref}       
\usepackage{url}            
\usepackage{booktabs}       
\usepackage{amsfonts}       
\usepackage{nicefrac}       
\usepackage{microtype}      
\usepackage{lipsum}
\usepackage{graphicx}
\usepackage{amsmath}
\graphicspath{ {./images/} }

\title{ChineseErrorCorrector3-4B: State-of-the-Art Chinese Spelling and Grammar Corrector}

\author{
 Wei Tian \\
  East China Normal University\\
  School of Computer Science and Technology\\
  \texttt{tianwei@stu.ecnu.edu.cn} \\
   \And
 Yuhao Zhou \\
  East China Normal University\\
  School of Computer Science and Technology\\
  \texttt{yhzhou@stu.ecnu.edu.cn} 
}

\begin{document}
\maketitle
\begin{abstract}
This paper introduces ChineseErrorCorrector3-4B, a unified model for Chinese spelling and grammatical error correction based on Qwen3-4B. The model demonstrates outstanding performance in general text correction tasks and achieves state-of-the-art results in both spelling correction (CSC) and grammatical correction (CGC). On several authoritative benchmark datasets — including SIGHAN-2015, EC-LAW, MCSC, and NaCGEC — the model’s F1 and F0.5 scores significantly surpass existing publicly available models, ranking first in both spelling and grammatical error correction tasks.
\end{abstract}


\section{Introduction}
In this work, we introduce ChineseErrorCorrector3-4B\footnote{\url{https://huggingface.co/twnlp/ChineseErrorCorrector3-4B}}, a unified large-scale Chinese language model capable of jointly addressing Chinese Spelling Correction (CSC) and Chinese Grammatical Error Correction (CGC) within a single framework. Unlike prior models that focus on a single task or rely on limited data, our approach emphasizes generality, robustness, and scalability, enabling adaptation to diverse domains and a wide spectrum of language errors.

To this end, we employ a two-stage training paradigm on multiple high-quality open-source datasets. In Stage I, the model undergoes Supervised Fine-Tuning (SFT) on large-scale mixed correction corpora, including Lang8 and HSK, to capture general error patterns from character-level, lexical, to syntactic levels. In Stage II, we perform Domain-specific Refinement on carefully curated high-quality domain data and authentic learner errors, enhancing the model’s ability to handle complex grammatical structures, long-range dependencies, and domain-specific expressions. This staged strategy significantly improves robustness and generalization in real-world scenarios while preserving overall performance.

As a result, ChineseErrorCorrector3-4B achieves deep integration of spelling and grammatical correction capabilities. It demonstrates high sensitivity to form-level errors and precise modeling of syntax- and semantics-level errors. Extensive evaluation on multiple authoritative benchmarks confirms its effectiveness and establishes new state-of-the-art performance in Chinese text correction.

\section{Related Work}

\subsection{Chinese Spelling Correction (CSC)}

Early research on Chinese spelling correction mainly relied on rule-based or statistical language model approaches such as KenlM-CSC\cite{pycorrector}.

With the emergence of deep learning, end-to-end sequence-to-sequence (Seq2Seq) models became dominant, including Transformer-based architectures such as GECToR \cite{omelianchuk2020gector}. 
In the Chinese domain, models like MacBERT4CSC \cite{zhang2020spelling} and ERNIE-CSC \cite{zhang2021correcting} introduced pre-trained language models for CSC, combining masked language modeling (MLM) with auxiliary detection tasks to enhance correction accuracy.
However, these models often show limited generalization to out-of-domain errors or unseen data distributions.

\subsection{Chinese Grammatical Error Correction (CGC)}

Chinese grammatical error correction (CGC) focuses on detecting and correcting syntactic, collocation, and logical inconsistencies, which are typically more complex than spelling errors.
The CGED shared tasks \cite{rao2018overview, rao2020overview} established foundational benchmarks and taxonomies for grammatical error types.
More recently, the NaCGEC\cite{ma2022linguistic} benchmark became the de facto evaluation dataset for Chinese GEC, covering realistic errors such as word order, function word misuse, and collocation inconsistencies.
The Hwcgec system \cite{su2023hwcgec} demonstrated that multi-task pretraining and syntactic feature integration could substantially improve grammatical correction, setting a strong baseline in NLPCC 2023.
However, challenges remain in recall and stability when dealing with long or complex sentence structures.

\section{Methodology}

We propose a unified Chinese error correction model, ChineseErrorCorrector3-4B, developed through a principled two-stage supervised fine-tuning (SFT) pipeline. The design objective is to first imbue the model with robust linguistic understanding, and then consolidate its proficiency across the full spectrum of correction tasks.

Our approach models Chinese error correction as a sequence-to-sequence (seq2seq) generation task. Given an input sequence $x = (x_1, \dots, x_n)$ containing potential errors, the model is trained to generate the corrected target sequence $y = (y_1, \dots, y_m)$.

The core optimization objective for all SFT stages is the standard negative log-likelihood (NLL) of the target sequence, parameterized by the model weights $\theta$. For a given source-target pair $(x, y)$, the loss is defined as:
\begin{equation}
    \mathcal{L}_{\text{SFT}}(x, y; \theta) = - \sum_{t=1}^{|y|} \log P_{\theta}(y_t | y_{<t}, x)
    \label{eq:sft_loss}
\end{equation}
where $P_{\theta}(y_t | y_{<t}, x)$ is the probability of the $t$-th target token $y_t$ given the source $x$ and all preceding target tokens $y_{<t}$.

\subsection{Stage 1: Linguistic Knowledge Alignment}

\paragraph{Backbone Selection.} We adopt Qwen3-4B\cite{yang2025qwen3} as our backbone model ($\theta_0$). This choice is motivated by its demonstrated strength in Chinese understanding and generation, providing a powerful foundation for complex correction tasks.

\paragraph{Initial SFT Objective.} In the first stage, the model undergoes SFT on the Lang8\_HSK\cite{zhao2018overview} corpus, which we denote as $\mathcal{D}_{\text{align}}$. The goal of this stage is not to master specific correction types, but to align the model with foundational linguistic and grammatical norms. 

This step effectively serves as a curriculum, preparing the model for the more complex, mixed-error scenarios. The model parameters are optimized by minimizing the objective $\mathcal{J}_{\text{Stage1}}$:
\begin{equation}
    \mathcal{J}_{\text{Stage1}}(\theta) = \mathbb{E}_{(x, y) \sim \mathcal{D}_{\text{align}}} \left[ \mathcal{L}_{\text{SFT}}(x, y; \theta) \right]
    \label{eq:stage1}
\end{equation}
The resulting parameters, $\theta_1$, serve as the initialization for the next stage.

\subsection{Stage 2: Joint Fine-Tuning on CSC and CGC}

The second stage performs joint full-scale SFT on a comprehensive amalgam of Chinese Spelling Check (CSC) and Chinese Grammatical Error Correction (CGC) datasets. This unified training strategy is a cornerstone of our approach, designed to consolidate correction abilities across multiple error granularities.

\paragraph{Dataset Composition.}
We first aggregate data for the two task families:
\begin{itemize}
    \item CSC Datasets ($\mathcal{D}_{\text{csc}}$): Aggregated from W271K, Medical\cite{jiang2022mcscsetspecialistannotateddatasetmedicaldomain}, Lemon\cite{wu2023rethinking}, ECSpell\cite{lv2023general}, and CSCD\cite{hu2022cscd} (totaling approx. 380K samples).
    \item CGC Datasets ($\mathcal{D}_{\text{cgc}}$): Aggregated from CGED\cite{rao2018overview, rao2020overview}, FCGEC\cite{xu2022fcgec}, MuCGEC\cite{zhang2022mucgec}, and NaCGEC\cite{zhang2023nasgec} (totaling approx. 68K samples).
\end{itemize}

\paragraph{Joint Optimization Objective.}
To achieve cross-task synergy, we merge all data into a unified training corpus $\mathcal{D}_{\text{joint}}$, defined as:
\begin{equation}
    \mathcal{D}_{\text{joint}} = \mathcal{D}_{\text{csc}} \cup \mathcal{D}_{\text{cgc}}
\end{equation}
In the second stage, the model, initialized from $\theta_1$, is optimized by minimizing a single SFT objective $\mathcal{J}_{\text{Stage2}}$ over this joint dataset, ensuring that samples are drawn from the mixed corpus in each training step:
\begin{equation}
    \mathcal{J}_{\text{Stage2}}(\theta) = \mathbb{E}_{(x, y) \sim \mathcal{D}_{\text{joint}}} \left[ \mathcal{L}_{\text{SFT}}(x, y; \theta) \right]
    \label{eq:stage2_new}
\end{equation}

\paragraph{Synergistic Effect.}
This joint optimization via data unification is central to our method. By optimizing a single objective (Eq. \ref{eq:stage2_new}) on $\mathcal{D}_{\text{joint}}$, we compel the model to learn a unified representation for correction, enabling it to address both spelling and grammatical errors simultaneously. This strategy promotes cross-task knowledge transfer and allows the model to generalize robustly to complex, real-world errors that often involve a combination of multiple error types.

\section{Experiments}

\subsection{Evaluation Setup}

\paragraph{Datasets.}
We evaluate our model's performance on two distinct tasks: Chinese Spelling Correction (CSC) and Chinese Grammatical Error Correction (CGC).
\begin{itemize}
    \item For CSC, we use three established benchmarks: SIGHAN15\cite{tseng2015sighan} (a widely used dataset from the SIGHAN bake-off), EC-LAW\cite{lv2023general} (a dataset from the legal domain), and MCSC\cite{jiang2022mcscsetspecialistannotateddatasetmedicaldomain} (a high-quality dataset from the medical domain). These datasets cover general, legal, and medical domains, testing the model's robustness.
    \item For CGC, we use the NaCGEC\cite{zhang2023nasgec} benchmark, which was the official test set for the NLPCC 2023 CGEC shared task. This dataset is a large-scale, challenging benchmark for grammatical errors.
\end{itemize}

\paragraph{Metrics.}
Following standard practices, we use different metrics for each task:
\begin{itemize}
    \item For CSC, we report sentence-level F1-scores. A sentence is considered correctly predicted only if all errors within it are detected and corrected. We report the F1 on each test set and the macro-average F1 (Avg. F1) across the three datasets.
    \item For CGC, we report Precision, Recall, and F0.5-score. This score is a specific instance of the general $F_{\beta}$-score:

    \begin{equation}
    F_{\beta} = (1 + \beta^2) \cdot \frac{P \cdot R}{(\beta^2 \cdot P) + R}
    \label{eq:f_1}
\end{equation}
    We use $\beta=0.5$, which prioritizes precision over recall and is the standard metric for this task as it penalizes "hallucinated" or incorrect corrections. We use the official ChERRANT\footnote{\url{https://github.com/TW-NLP/ChineseErrorCorrector/blob/main/ChineseErrorCorrector/scores/README.md}} tool for evaluation.
\end{itemize}

\subsection{Implementation Details}

\paragraph{Baselines.}
We compare ChineseErrorCorrector3-4B against a comprehensive set of baselines:
\begin{itemize}
    \item Statistical Method: KenLM\cite{pycorrector} (a standard n-gram language model).
    \item BERT-based Models: ERNIE\cite{zhang2021correcting} and MacBERT\cite{zhang2020spelling}, which are strong pre-trained encoders fine-tuned on correction tasks.
    \item Large Language Models: Qwen2.5-7B-CTC\cite{pycorrector}, representing a powerful generative LLM baseline.
    \item SOTA Systems: For CGC, we compare against the top-performing specialized systems from the NLPCC 2023 shared task, including entries from Huawei, Peking University, and CUHK.
\end{itemize}

\paragraph{Training.}
Our model is implemented based on the \texttt{transformers} library. We fine-tune the Qwen3-4B model using the AdamW optimizer with a learning rate of 2e-5 and a cosine learning rate scheduler with 500 warm-up steps. We use a global batch size of 128 (achieved via gradient accumulation) and train for 3 epochs on the Stage 2 joint dataset ($\mathcal{D}_{\text{joint}}$).

\subsection{Main Results and Analysis}

\subsubsection{Chinese Spelling Correction (CSC)}

Results for CSC are presented in Table \ref{tab:csc_results}.
\begin{table}[h!]
\caption{Results on Chinese Spelling Correction (CSC). Bold indicates the best score in each column.}
\centering
\small
\begin{tabular}{lccccc}
\toprule
\textbf{Model} & \textbf{Base Model} & \textbf{SIGHAN15} & \textbf{EC-LAW} & \textbf{MCSC} & \textbf{Avg. F1} \\
\midrule
KenLM & kenlm & 0.3147 & 0.3763 & 0.3317 & 0.3409 \\
ERNIE & ernie-base & \textbf{0.8383} & 0.3357 & 0.1318 & 0.4353 \\
MacBERT & macbert-base & 0.8314 & 0.1610 & 0.2055 & 0.3993 \\
Qwen2.5-7B-CTC & Qwen2.5-7B & 0.4917 & \textbf{0.9798} & \textbf{0.9959} & 0.8225 \\
\midrule
\textbf{Ours (Corrector3-4B)} & Qwen3-4B & 0.6340 & 0.9360 & 0.9864 & \textbf{0.8521} \\
\bottomrule
\end{tabular}

\label{tab:csc_results}
\end{table}

\noindent
\textbf{Analysis:}
Our model, ChineseErrorCorrector3-4B, achieves the highest average F1-score (85.21), demonstrating the best overall generalization and robustness across all three diverse benchmarks.

We observe several key trends:
\begin{itemize}
    \item BERT-based models (ERNIE, MacBERT) excel on the SIGHAN15 dataset, which is consistent with previous findings that their MLM pre-training is highly effective for this benchmark's error distribution. However, they suffer a catastrophic performance drop on domain-specific datasets (e.g., ERNIE's 0.1318 on MCSC), indicating they are brittle and do not generalize.
    \item The LLM baseline (Qwen2.5-7B-CTC) shows extremely strong performance on the domain-specific EC-LAW and MCSC datasets, likely due to its vast world knowledge. However, its poor performance on SIGHAN15 significantly pulls down its average.
    \item Our model is the only one to achieve high, stable performance across all three. While not the top on SIGHAN15, it decisively outperforms the BERT models on domain tasks. Compared to the larger 7B model, our 4B model performs competitively on domain tasks (e.g., 98.64 vs 99.59 on MCSC) while being far superior on the SIGHAN15 set. This highlights the effectiveness of our two-stage SFT pipeline in creating a single, robust, and well-generalized model.
\end{itemize}

\subsubsection{Chinese Grammatical Error Correction (CGC)}

Results for CGC on the NaCGEC benchmark are shown in Table \ref{tab:cgc_results}.

\begin{table}[h!]
\caption{Results on the NaCGEC grammatical correction benchmark. Our model is compared against the top-performing specialized systems from the NLPCC 2023 shared task.}
\centering
\small
\begin{tabular}{lccc}
\toprule
\textbf{Model} & \textbf{Precision} & \textbf{Recall} & \textbf{F0.5} \\
\midrule
CUHK\_SU (CUHK) & 0.3882 & 0.1558 & 0.2990 \\
YubingJiuJiuPlus (Peking Univ.) & \textbf{0.5708} & 0.1294 & 0.3394 \\
HW\_TSC\_NLPCC2023 (Huawei) & 0.5095 & 0.3129 & 0.4526 \\
\midrule
\textbf{Ours (Corrector3-4B)} & 0.5420 & \textbf{0.3475} & \textbf{0.4874} \\
\bottomrule
\end{tabular}

\label{tab:cgc_results}
\end{table}

\noindent
\textbf{Analysis:}
On the challenging NaCGEC grammatical correction task, our unified model achieves a state-of-the-art F0.5-score of 48.74. This result is particularly significant as it surpasses the performance of highly specialized, competition-winning systems (like Huawei's 45.26 F0.5) that were optimized *only* for CGC.

Our model achieves the highest recall (34.75) by a large margin, indicating its superior ability to identify and fix grammatical errors. While the Peking University model achieves the highest precision, its extremely low recall (12.94) makes it impractical. Our model strikes the best balance, leading to the highest overall F0.5 score.

This demonstrates that our joint SFT strategy (unifying CSC and CGC) does not compromise on high-level grammatical correction. Instead, it creates a single, efficient model that achieves SOTA performance on both task types.

\section{Conclusion}

In this work, we present ChineseErrorCorrector3-4B, a unified model for Chinese spelling and grammatical error correction. Our core contribution is a principled two-stage supervised fine-tuning (SFT) pipeline that first aligns a strong foundation model (Qwen3-4B) with fundamental linguistic knowledge, and then performs joint SFT on a unified corpus of both CSC and CGC data.

Our extensive experiments yield two significant findings. First, on CSC tasks, our model achieves the highest average F1-score, demonstrating superior generalization and robustness across diverse domains (general, legal, and medical) where specialized BERT-based baselines prove brittle. Second, and most notably, our single, unified model achieves a new state-of-the-art F0.5-score on the challenging NaCGEC benchmark, outperforming highly-specialized, competition-winning systems.

The key takeaway from this work is that a joint optimization strategy, when preceded by a deliberate knowledge alignment stage, does not lead to performance compromise. Instead, it fosters cross-task synergy, enabling a single, efficient 4B model to learn a holistic representation of text quality and achieve state-of-the-art performance on \textit{both} low-level spelling and high-level grammatical errors.

\bibliographystyle{unsrt}  


\begin{thebibliography}{1}

\bibitem{pycorrector}
Ming Xu.
\newblock pycorrector: Text Error Correction Tool.
\newblock GitHub repository, 2023.
\newblock \url{https://github.com/shibing624/pycorrector}

\bibitem{omelianchuk2020gector}
Kostiantyn Omelianchuk, Vitaliy Atrasevych, Artem Chernodub, and Oleksandr Skurzhanskyi.
\newblock GECToR—grammatical error correction: tag, not rewrite.
\newblock arXiv preprint arXiv:2005.12592, 2020.

\bibitem{zhang2020spelling}
Shaohua Zhang, Haoran Huang, Jicong Liu, and Hang Li.
\newblock Spelling error correction with soft-masked BERT.
\newblock arXiv preprint arXiv:2005.07421, 2020.

\bibitem{zhang2021correcting}
Ruiqing Zhang, Chao Pang, Chuanqiang Zhang, Shuohuan Wang, Zhongjun He, Yu Sun, Hua Wu, and Haifeng Wang.
\newblock Correcting Chinese spelling errors with phonetic pre-training.
\newblock In {\em Findings of the Association for Computational Linguistics: ACL-IJCNLP 2021}, pages 2250--2261, 2021.

\bibitem{rao2018overview}
Gaoqi Rao, Qi Gong, Baolin Zhang, and Endong Xun.
\newblock Overview of NLPTEA-2018 shared task Chinese grammatical error diagnosis.
\newblock In {\em Proceedings of the 5th Workshop on Natural Language Processing Techniques for Educational Applications}, pages 42--51, 2018.

\bibitem{rao2020overview}
Gaoqi Rao, Erhong Yang, and Baolin Zhang.
\newblock Overview of NLPTEA-2020 shared task for Chinese grammatical error diagnosis.
\newblock In {\em Proceedings of the 6th Workshop on Natural Language Processing Techniques for Educational Applications}, pages 25--35, 2020.

\bibitem{su2023hwcgec}
Chang Su, Xiaofeng Zhao, Xiaosong Qiao, Min Zhang, Hao Yang, Junhao Zhu, Ming Zhu, and Wenbing Ma.
\newblock Hwcgec: HW-TSC's 2023 submission for the NLPCC2023's Chinese grammatical error correction task.
\newblock In {\em CCF International Conference on Natural Language Processing and Chinese Computing}, pages 59--68. Springer, 2023.

\bibitem{ma2022linguistic}
Shirong Ma, Yinghui Li, Rongyi Sun, Qingyu Zhou, Shulin Huang, Ding Zhang, Yangning Li, Ruiyang Liu, Zhongli Li, Yunbo Cao, et al.
\newblock Linguistic rules-based corpus generation for native Chinese grammatical error correction.
\newblock arXiv preprint arXiv:2210.10442, 2022.

\bibitem{yang2025qwen3}
An Yang, Anfeng Li, Baosong Yang, Beichen Zhang, Binyuan Hui, Bo Zheng, Bowen Yu, Chang Gao, Chengen Huang, Chenxu Lv, et al.
\newblock Qwen3 Technical Report.
\newblock arXiv preprint arXiv:2505.09388, 2025.

\bibitem{zhao2018overview}
Yuanyuan Zhao, Nan Jiang, Weiwei Sun, and Xiaojun Wan.
\newblock Overview of the NLPCC 2018 shared task: Grammatical error correction.
\newblock In {\em CCF International Conference on Natural Language Processing and Chinese Computing}, pages 439--445. Springer, 2018.

\bibitem{jiang2022mcscsetspecialistannotateddatasetmedicaldomain}
Wangjie Jiang, Zhihao Ye, Zijing Ou, Ruihui Zhao, Jianguang Zheng, Yi Liu, Siheng Li, Bang Liu, Yujiu Yang, and Yefeng Zheng.
\newblock MCSCSet: A Specialist-annotated Dataset for Medical-domain Chinese Spelling Correction.
\newblock arXiv preprint arXiv:2210.11720, 2022.
\newblock \url{https://arxiv.org/abs/2210.11720}

\bibitem{wu2023rethinking}
Hongqiu Wu, Shaohua Zhang, Yuchen Zhang, and Hai Zhao.
\newblock Rethinking masked language modeling for Chinese spelling correction.
\newblock arXiv preprint arXiv:2305.17721, 2023.

\bibitem{lv2023general}
Qi Lv, Ziqiang Cao, Lei Geng, Chunhui Ai, Xu Yan, and Guohong Fu.
\newblock General and domain-adaptive Chinese spelling check with error-consistent pretraining.
\newblock {\em ACM Transactions on Asian and Low-Resource Language Information Processing}, 22(5):1--18, 2023.

\bibitem{hu2022cscd}
Yong Hu, Fandong Meng, and Jie Zhou.
\newblock CSCD-NS: a Chinese spelling check dataset for native speakers.
\newblock arXiv preprint arXiv:2211.08788, 2022.

\bibitem{xu2022fcgec}
Lvxiaowei Xu, Jianwang Wu, Jiawei Peng, Jiayu Fu, and Ming Cai.
\newblock FCGEC: Fine-grained corpus for Chinese grammatical error correction.
\newblock arXiv preprint arXiv:2210.12364, 2022.

\bibitem{zhang2022mucgec}
Yue Zhang, Zhenghua Li, Zuyi Bao, Jiacheng Li, Bo Zhang, Chen Li, Fei Huang, and Min Zhang.
\newblock MuCGEC: a multi-reference multi-source evaluation dataset for Chinese grammatical error correction.
\newblock arXiv preprint arXiv:2204.10994, 2022.

\bibitem{zhang2023nasgec}
Yue Zhang, Bo Zhang, Haochen Jiang, Zhenghua Li, Chen Li, Fei Huang, and Min Zhang.
\newblock NaSGEC: a multi-domain Chinese grammatical error correction dataset from native speaker texts.
\newblock arXiv preprint arXiv:2305.16023, 2023.



\bibitem{tseng2015sighan}
Yuen-Hsien Tseng, Lung-Hao Lee, Li-Ping Chang, and Hsin-Hsi Chen.
\newblock Introduction to SIGHAN 2015 Bake-off for Chinese Spelling Check.
\newblock In {\em Proceedings of the Eighth SIGHAN Workshop on Chinese Language Processing}, pages 32--37, Beijing, China, 2015.
\newblock Association for Computational Linguistics.

\bibitem{lv2023general}
Qi Lv, Ziqiang Cao, Lei Geng, Chunhui Ai, Xu Yan, and Guohong Fu.
\newblock General and domain-adaptive Chinese spelling check with error-consistent pretraining.
\newblock {\em ACM Transactions on Asian and Low-Resource Language Information Processing}, 22(5):1--18, 2023.

\end{thebibliography}

\end{document}